# AI and Consumer Rights in India
## Working Paper

**Omir Kumar**
Policy Analyst
Centre for Responsible AI
IIT Madras
Chennai, India

**Sriya Sridhar**
Senior Policy Analyst
Centre for Responsible AI
IIT Madras
Chennai, India

**Vibhav Mithal**
Associate Research Fellow
Centre for Responsible AI
IIT Madras
Chennai, India

**Balaraman Ravindran**
Head, Wadhwani School of Data Science & AI
Centre for Responsible AI
IIT Madras
Chennai, India

## ABSTRACT

As AI systems proliferate in consumer-facing applications, questions about liability for AI-related harms remain unresolved. This working paper examines whether India's Consumer Protection Act, 2019, adequately addresses harm caused by defective AI products and services, and whether it proportionately allocates liability across the AI value chain.

The Act's broad definitions of product liability, harm, and deficiency appear technology-agnostic and potentially applicable to AI-related incidents—including personal injury, psychological harm, biased outputs, and loss of control. However, significant gaps remain. Proving causation between AI defects and consumer harm presents a technical challenge, as AI failures often stem from design choices rather than discrete defects. Additionally, the Act's framework assumes distinct roles for manufacturers, sellers, and service providers, yet the AI value chain involves overlapping responsibilities among data providers, model developers, deployers, and users that do not neatly map to these categories. Current liability frameworks lack proportionate mechanisms to effectively address complex, multi-stakeholder AI harms. While the Act may cover AI entities, enforcement requires clarification on sector-specific overlaps.

## 1. Introduction

As consumers, whether knowingly or unknowingly, we are interacting with some form of AI. From the use of AI chatbots to the integration of AI into businesses' supply chains, the spectrum of consumer-facing AI applications is wide. In all such applications, when the technology works as desired, the results can be remarkably efficient. However, there are instances when the outcomes are not desirable and may also cause harm to consumer autonomy and well-being.[1] These risks include invasion of privacy, biased outcomes, and unreliability. From the perspective of consumers, who is responsible for the harms caused by AI-enabled applications, and is there any legal remedy available?

India's Consumer Protection Act, 2019 (Act), provides consumers with certain rights and protections.[2] While repealing the Consumer Protection Act, 1986, the 2019 Act codified the concept of product liability for the first time. Product liability places legal responsibility for harm, for any damage caused by a product/service, on *all parties* in the supply chain. The legal concept acknowledges a value chain and the aspect that liability/responsibility for harm falls upon all actors within a particular value chain. Product liability may also be viewed as *a* legal encapsulation of the AI Governance Guidelines' accountability recommendation of a "graded liability regime" and "transparency in AI value chain".[3]

Under the Consumer Protection Act, for a product liability action, consumers can claim compensation for harms caused by defective products or services. The Act defines three types of responsible parties: (1) manufacturers, (2) sellers, and (3) service providers. It also lays down the liability for all these entities.

It is unclear whether the product liability provision adequately and proportionately covers AI-related harm and entities in the AI value chain (which may or may not include "manufacturer", "seller" or "service provider" and may also be termed differently). Given the increasing uptake of AI products and/or services by both consumers and enterprises, and the risk of harm *consequent thereto*, it is important to study the product liability provision of the Act to assess if the same may be invoked for AI products and/or services. This is also in line with the recommendation in the India AI Governance Guidelines ('the Guidelines'), which calls for using existing laws to govern AI applications.[4] While the Guidelines talk about applications (for example, chatbots), we will be including AI models (the underlying technology powering AI applications) in our analysis.

In the first paper of our series on AI and consumer rights in India, we seek to answer two questions. Does the Act cover AI-related harm and accord proportionate liability to different actors in the AI supply chain?

## 2. AI-related harm covered by the Consumer Protection Act

[1] Paterson JM, Maker Y. Consumer Protection Law and AI. In: Lim E, Morgan P, eds. The Cambridge Handbook of Private Law and Artificial Intelligence. Cambridge Law Handbooks. Cambridge University Press; 2024:113–134.

[2] Consumer Protection Act, 2019, Government of India, https://www.indiacode.nic.in/bitstream/123456789/15256/1/eng201935.pdf

[3] India AI Governance Guidelines, Government of India, https://static.pib.gov.in/WriteReadData/specificdocs/documents/2025/nov/doc2025115685601.pdf

[4] Ibid.

The Act covers a wide range of services, including banking, financing, insurance, transportation, and entertainment. The definition is wide enough to cover AI-enabled applications (such as chatbots offering medical advice or wellness services). However, the Act does not include the rendering of any service free of charge.[5] So this may exclude incidents that occur while using free AI-based applications.

The Act defines product liability as the responsibility of a product manufacturer, product service provider, or product seller to compensate for harm caused to a consumer by such defective product/deficient service. The Act lays down certain grounds, which may make (1) a manufacturer, or (2) product service provider, or (3) product seller liable for harm to consumers. The grounds include, among others, (a) services or products not conforming to the terms & conditions, (b) inadequate instructions for correct usage to prevent harm. Harm is defined broadly under the Act. It includes (i) damage to any property, other than the product itself; (ii) personal injury, illness, or death; (iii) mental agony or emotional distress attendant to personal injury or illness or damage to property. It does not include damage to the property on account of breach of warranty conditions or any commercial or economic loss, including any direct, incidental, or consequential loss. As product liability extends to deficiency in services, the Act also defines deficiency. Deficiency includes fault or imperfection in the service, and also includes negligence, which causes loss or injury to the consumer. The definitions of harm and deficiency are technology-agnostic and wide enough to cover AI-related incidents. The following table groups AI-related harm in three broad categories. Further, it lays out different grounds provided in the Act that may cover that harm. The table is merely an illustration to demonstrate the applicability of the product liability provisions on AI harm. This is subject to interpretation by enforcement authorities and the judiciary.

**Table 1: Illustrative mapping of AI-related harm to grounds of harm and deficiency under the Consumer Protection Act, 2019.**

| **Type of harm** | **Example** | **Grounds of harm & deficiency in the Act that may cover the harm** |
|---|---|---|
| Personal injury to consumers | Chatbots encouraging self-harm | personal injury, illness, or death; mental agony or emotional distress attendant to personal injury or illness or damage to property |
| Inaccurate/false outputs | Customers getting false information about a refund policy from a chatbot | Deficiency in services |
| Loss of control | Coding agent wiping out a database without user authorisation, or an autonomous car overriding user instructions | Deficiency in services, personal injury, damage to property |

[5] Section 2(42), Consumer Protection Act, 2019, Government of India

### 3. Establishing the actual/proximate cause for the AI-related harm

Even assuming that AI harms are covered by the Act, the next step is to demonstrate that the defective product or deficient/faulty/imperfect/inadequate service was the actual or proximate cause of harm. Simply put, one needs to prove the causal link between the harm and the AI-related activity that allegedly led to the harm. For example, how do we establish that the responses of a chatbot motivated a person to engage in behaviour that may be harmful to themselves and/or others? Testing for 'defectiveness' and a 'causal' link in the way that consumer protection regulators may be able to do for more traditional products or services will be difficult in this context.

A recent ruling in the US provides a useful framework for product liability design-defect to look at such cases.[6] A jury in California found Meta and Google liable for building addictive social media platforms that harmed the mental health of a woman who used them as a child.[7] The plaintiff argued that design features like algorithmic recommendations, autoplay, and likes were made to maximise engagement, and that this addiction fuelled her mental health issues.[8] Several technical papers show how AI hallucinations are a feature of AI design and not a bug,[9] [10] which raises the question of how authorities might scrutinise the *design* of AI-driven technologies in the future, rather than the *effects*. However, it is important to distinguish such a product liability action in the USA from the regime in India, which tends to focus on the processes of contracting for and the performance of consumer products.

In the context of the design of consumer-facing technological interfaces, India's Central Consumer Protection Authority (CCPA) has notified Guidelines prohibiting dark patterns on digital platforms, banning certain types of dark patterns that have the effect of undermining consumer decision-making or autonomy.[11] Recently, the CCPA also issued an enforcement action in this regard against misleading advertisements and default platform features which mislead consumers into making decisions that they otherwise might not have - demonstrating a growing trend towards authorities acting *before* harm takes place to consumers rather than remedies *after* the fact.[12] As AI is integrated into these interfaces, for example, in relation to advertising, pricing algorithms, and potential coercion by chatbots, there may be merit in revisiting these guidelines and collecting more evidence on the way that AI may create new dark patterns or exacerbate existing ones.

---

[6] "Technology landmark US case could launch a whole wave of addiction litigation", International Bar Association, March 2026, https://www.ibanet.org/Technology-landmark-US-case-could-launch-a-whole-wave-of-addiction-litigation

[7] *P. F., et al. (K.G.M.) v. Meta Platforms, Inc., et al.* See also, "Jury finds Instagram and YouTube liable in a landmark social media addiction trial", The Associated Press, March 2026, https://apnews.com/article/social-media-addiction-trial-la-5e54075023d837ccdc76c4ca512e925d

[8] According to news reports, the companies will appeal this decision.

[9] Zhao, Wenting, et al. "WildHallucinations: Evaluating Long-form Factuality in LLMs with Real-World Entity Queries." ArXiv, 2024, https://arxiv.org/abs/2407.17468

[10] Chandra, Kartik, et al. "Sycophantic Chatbots Cause Delusional Spiraling, Even in Ideal Bayesians." ArXiv, 2026, https://arxiv.org/abs/2602.19141

[11] Guidelines for Prevention and Regulation of Dark Patterns, 2023, Central Consumer Protection Authority.

[12] Central Consumer Protection Authority, Krishi Bhawan, New Delhi – 110001, Case No. CCPA-2/94/2025-CCPA, https://jagograhakjago.gov.in/CCPA_Orders/Physics_Wallah_Limited_Order_01June2026.pdf

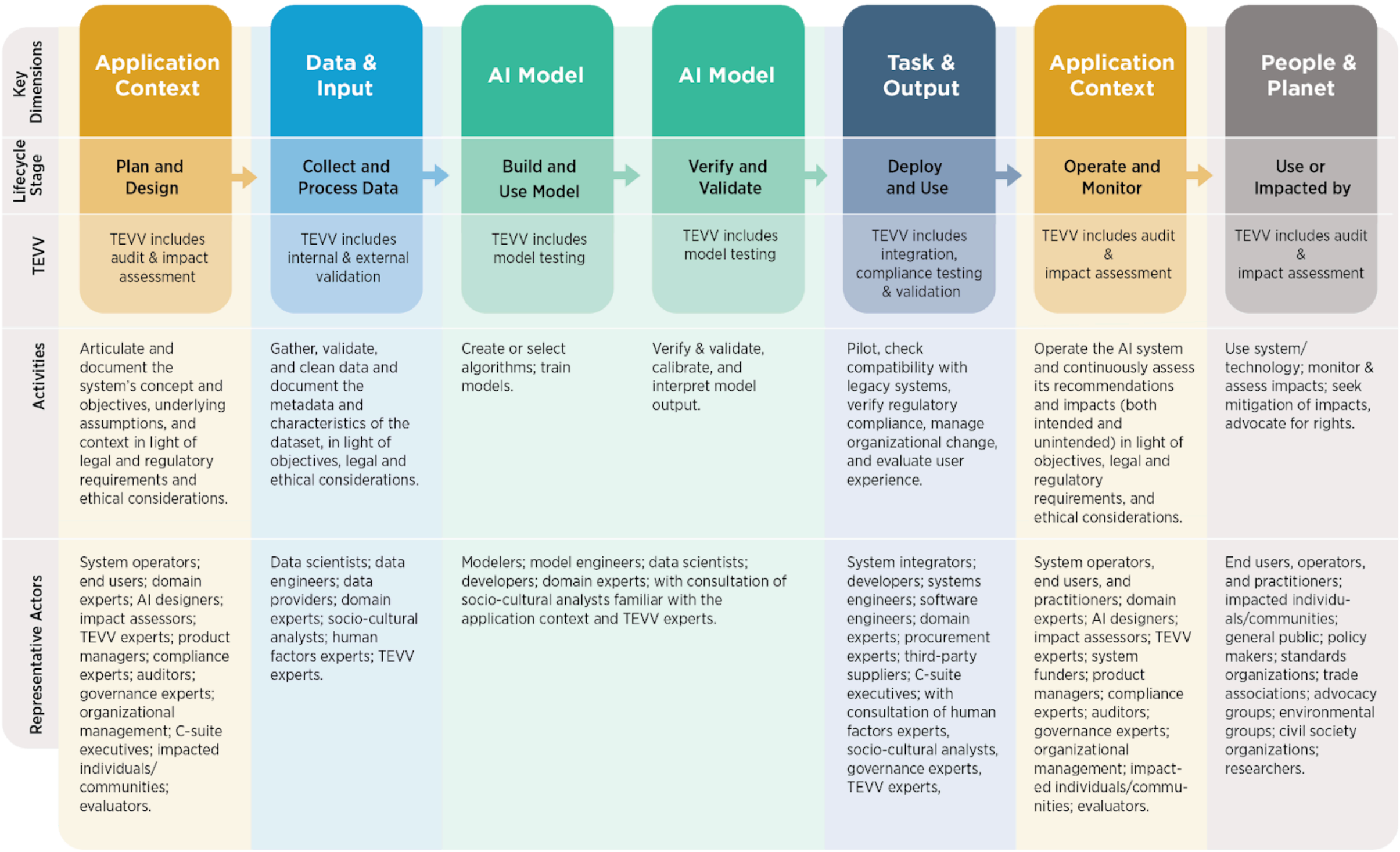

**Figure 1. Source:** *AI Risk Management Framework (AI RMF), National Institute of Standards and Technology*

### 4. Attributing liability

You apply for a loan, and an AI system rejects you because it was trained on biased data. Who should be held responsible? The bank that used the AI? The company that built the model? Or the organisation that provided the flawed data? In the context of liability and AI harm, the RBI FREE-AI Committee (2025) noted that AI deployments blur the lines of responsibility between various stakeholders.[13] It noted that this may lead to institutions facing risks (such as legal risk and regulatory sanctions), especially in instances where AI-driven decisions impact customer rights, credit approvals, or investment outcomes. For instance, recently, an AI coding agent powered by Claude deleted an entire company's database and all its backups, impacting all its consumers.[14] Who is liable? The AI company? The coding tool that deployed it? Or the company that gave the AI access? The challenge is that AI isn't created by just one entity. It involves a chain of players. Based on Figure 1 of the NIST framework, there are broadly three entities in the AI value chain: developer, deployer, and user.

[13] Reserve Bank of India, *Report of the FREE-AI Committee (2025)*, August 2025, https://rbidocs.rbi.org.in/rdocs/PublicationReport/Pdfs/FREEAIR130820250A24FF2D4578453F824C72ED9F5D5851.PDF

[14] "Claude-powered AI agent's confession after deleting a firm's entire database: 'I violated every principle I was given'", The Guardian, April 2026, https://www.theguardian.com/technology/2026/apr/29/claude-ai-deletes-firm-database

The Consumer Protection Act defines three types of entities:

- manufacturers (those who make a product)
- sellers (those who distribute or sell it)
- service providers (those who provide services using it).

Further, product means "any article, goods, substance, raw material, or extended cycle of such product." However, the Act doesn't specifically address AI or explain how liability should be divided among these different players. But given how wide these definitions are, it is likely that they can cover AI models and AI-based applications. Based on these definitions, data providers and model developers could both be considered "manufacturers" since they're involved in creating the AI product. Entities fine-tuning such models for different use cases may be covered as “sellers,” and if they are themselves providing the services, then also under “service providers.” Lastly, companies merely using fine-tuned models to provide services would come under “service providers”. Therefore, technically, the Act seems to cover everyone in the AI chain.

However, the challenge would be to see if the framework of product liability provided by the Act aligns with the AI value chain. For instance, a hospital uses an AI diagnostic tool that misidentifies a patient's cancer as benign, leading to delayed treatment and serious health consequences. Should the AI company that created the diagnostic algorithm be liable? Should the medical device manufacturer that integrated the AI into their equipment be responsible? Should the hospital that chose to rely on the AI's recommendation bear the blame? Or should the data company that provided the medical images used to train the model be held accountable if those images were unrepresentative of diverse patient populations? Maybe all of them, in varying degrees, but what framework do we use to proportionately accord liability?

The Act’s framework for product liability lays down grounds for consumers against which they can hold different entities accountable for harm caused. For instance, a product manufacturer can be held liable if the product contains a manufacturing defect or does not contain adequate instructions for use. Essentially, these are the responsibilities given to different entities to avoid harm to consumers. The underlying assumption in such a framework is that the role and value added by different entities in a product/service can be defined and distinguished. This is not the case with AI, where responsibility is fragmented and overlapping. The legal doctrine of chain-of-distribution liability in the USA can be a useful approach to thinking about AI liability.[15,16] This approach imposes liability on every party that played a part in the consumer getting the product.

---

[15] Vandermark v. Ford Motor Co., case explainer, Studicata, https://www.studicata.com/case-briefs/case/vandermark-v-ford-motor-co

[16] “Understanding Chain-of-Distribution Liability”, Weycer Law Firm, February 2024, https://weycerlawfirm.com/blog/understanding-chain-of-distribution-liability/. See also, Bryan H Choi, Negligence Liability for AI Developers, Lawfare, September 2024, https://www.lawfaremedia.org/article/negligence-liability-for-ai-developers.

## 5. Way forward

Broadly speaking, the Act seems adequate to handle AI-related cases, insofar as these systems are consumer-facing. However, it also has some gaps. The applicability and gaps could be studied by the recently constituted AI Governance and Economic Group and Technology and Policy Expert Committee, along with the Central Consumer Protection Authority (CCPA). They should clarify any overlaps with other sector-specific laws (for example, data protection law) and accordingly give recommendations. Apart from issues relating to the applicability of the Act, there is also the problem of institutional and technical capacity in law enforcement agencies and consumer awareness. These two factors may be the reason why we have not seen any AI-related consumer cases in courts. This is in sharp contrast to the enforcement of personality rights cases, where there is frequent litigation and action by Courts. To address this gap with respect to harms under the Consumer Protection Act, the AI Safety Institute should work on building the technical capacity of consumer forums, and the CCPA should work on spreading awareness around AI and consumer protection. Ultimately, we may also need to explore the need for complementary legal regimes to address the varied risks posed by AI systems integrated within consumer-facing products and services. This will also vary depending on context, the level of personalisation involved, and personal information collected, and individual/societal ramifications of any bias or decision-making. In some contexts, responses to the relationship between AI and consumers will need to be informed by policy and values around the ethical and fair use of AI. We aim to explore these issues in future working papers in this series.

**Acknowledgments**

This working paper is published by the Centre for Responsible AI (CeRAI), Indian Institute of Technology Madras, Chennai – 600036, Tamil Nadu, India (https://cerai.iitm.ac.in/). CeRAI is a multi-disciplinary, non-profit research centre positioned in the Global South, as one among the few global institutions that specialises in both technical and policy research to ensure and enable the responsible development and deployment of AI systems.

**Recommended citation:** Kumar, O., Sridhar, S., Mithal, V., & Ravindran, B. (2026, August). AI and Consumer Rights in India—Working Paper Series: Part 1. Centre for Responsible AI, Indian Institute of Technology Madras.

Errors in the paper, if any, are the sole responsibility of the authors. The copyright in this work vests with CeRAI.